# Full interpretable machine learning in 2D with inline coordinates

Boris Kovalerchuk  
*Dept. of Computer Science*  
*Central Washington University*  
USA  
BorisK@cwu.edu

Hoang Phan  
*Dept. of Computer Science*  
*Central Washington University*  
USA  
Hoang.Phan@cwu.edu

*Abstract*—This paper proposed a new methodology for machine learning in 2-dimensional space (2-D ML) in inline coordinates. It is a full machine learning approach that does not require to deal with n-dimensional data in n-dimensional space. It allows discovering n-D patterns in 2-D space without loss of n-D information using graph representations of n-D data in 2-D. Specifically, it can be done with the inline based coordinates in different modifications, including static and dynamic ones. The classification and regression algorithms based on these inline coordinates were introduced. A successful case study based on a benchmark data demonstrated the feasibility of the approach. This approach helps to consolidate further a whole new area of full 2-D *machine learning as a promising ML methodology*. It has advantages of abilities to involve actively the end-users into the discovering of models and their justification. Another advantage is providing interpretable ML models.

## I. Introduction

Interpretable Machine Learning (ML) is a major focus in Machine Learning domain these days [7,13,19]. The approaches range from explaining black box models to building explainable models from scratch. One of the attractive options is building explainable machine learning models using visual means. However, it is challenging, because data in machine learning are multidimensional, which we cannot see by the naked eye. So, the efficient tools are needed.

Traditional methods, which convert n-D data to 2-D data are lossy, not preserving all multidimensional information. In contrast the representation of n-D data using *General Line Coordinates* (GLC) is lossless [11]. This visual representation opened the opportunity for *full multidimensional machine learning in two dimensions without loss of information*.

The advantage of this methodology is two-fold. In simple situations it allows discovering the pattern visually just by observing these data visualized in GLC. In more complex situations, which are common in ML, it allows discovering patterns in 2-D representations using new 2-D ML methods. It is growing into a whole new field of machine learning. This paper is in this realm with the focus on a specific type of general line coordinates that are the inline coordinates [11].

In the series of our prior work [5, 12, 10, 14, 20] we demonstrated feasibility of full 2-D ML with other types of General Line Coordinates (Shifted Paired Coordinates, Elliptic Paired Coordinates, CPC-R and GLC-L). The studies in this realm can be traced to [3, 4] for Parallel Coordinates and [17] for 2-D modeling of non-image data. Often 2-D studies in ML cover only simple 2-D examples to illustrated ML algorithms visually. Next, visual analytics studies have been very active in exploring parallel coordinates for tasks related to clustering [16], but much less for supervised learning, which is the focus of our study. The studies in this area include [1, 18, 21].

We believe that it's time to consolidate all such studies within a general concept, which we call a **full 2D ML methodology**. Traditionally 2D studies in machine learning were considered as only an auxiliary exploratory data/model visualization with loss of n-D information mostly afterwards or before the actual machine learning. It was assumed that in 2-D we are losing n-D information, and we need full n-dimensional analysis in n-D space to construct ML models. The full 2-D ML methodology shows that it's not necessary. This methodology goes beyond visual knowledge discovery, which we advocated in [11]. It expands the studies from visual discovery by a human supported by ML methods, to a full scope of machine learning methods, for discovering full patterns *analytically* in 2-D.

Section II describes main concepts, Section III presents algorithms and Section IV covers the case study with a benchmark Wisconsin Breast Cancer (WBC) data from UCI ML repository [6] that demonstrates the feasibility of the approach. Section V presents conclusion and future studies including maturing of the proposed techniques.

## II. Definitions: In-Line Based Coordinates

The General Line Coordinates allow drawing *n* coordinate axes in 2-D in a variety of ways: curved, parallel, unparalleled, collocated, disconnected, and so on [11]. GLCs include **In-Line Coordinates** (ILC) shown in Fig. 1, which are similar to Parallel Coordinates, except that the axes $X_1, X_2, \ldots X_n$ are horizontal, not vertical. All coordinates are collocated on the same line, and may or may not overlap. A sequence of directed curves or polylines satisfies the requirement of lossless representation of n-D point in 2-D. The curves/polylines of different heights and shapes can show additional information such as the distance between adjacent values, $|x_i - x_{i+1}|$ like it is done in Fig. 1 [11].

In-Line Coordinates require the same number of nodes and links as Parallel Coordinates, which makes the scopes of applicability of these methods similar.

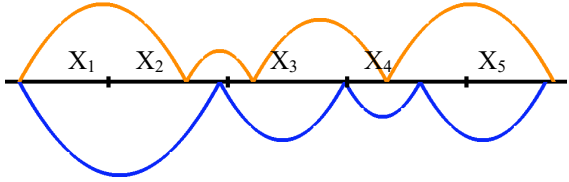

Fig. 1. Two 5-D points of two classes in In-Line coordinates.

The links between the nodes are directed edges, but we omit arrowheads when the direction is clear. To observe better the difference between n-D points of different classes, we can draw n-D points of one class above the coordinate line and n-D points of another class below (see Fig. 1). We consider the following *location modes* of ILC:

(L1) *Sequential* ILC with coordinates located one after another (Fig. 1).

(L2) *Collocated* ILC with coordinates drawn at the same location with full overlap (Fig. 2a).

(L3) *Generic* ILC where some coordinates can be *sequential*, *collocated*, *overlapping,* or *disjoined* (Fig. 2b).

(L4) *Dynamic* ILC with coordinated located dynamically as it is explained later.

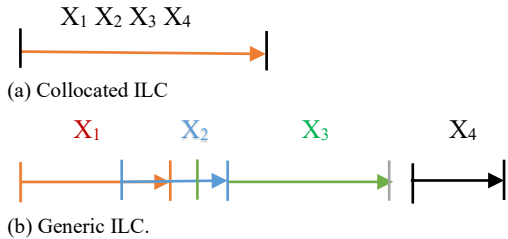

Fig. 2. Options to locate coordinates in on-line coordinates

In L3, a given n-D point **c** can be collapsed to a single point on ILC by selecting a specific ILC overlap [11]. It is a useful visual property when n-D point **c** is a center of the class with other n-D points of this class concentrated next to it. *Reordering coordinates* $X_1$-$X_n$ is another option, to make the patterns of interest more visible.

We have several options to **construct links** that connect points on $X_1$-$X_n$ by assigning meaning to their characteristics such as width and height to convey additional information. See an example in Fig. 3a for a 7-D point with values of $x_3$ and $x_4$ encoded as the **height** and the **width** of the line that connects $(x_1,x_2)$, and values of $x_6$ and $x_7$ the height and width of the line that connects $(x_2,x_5)$. Here, only three coordinates $X_1$, $X_2$ and $X_5$ are directly encoded in the baseline of ILC making it shorter. Fig. 3b and Fig. 4 show other options. Fig. 3b uses the lengths of **sides** of the line that connects points $x_1$ and $x_2$ to encode values of $x_3$ and $x_4$, instead of using its width and height. Similarly, lengths of sides of the line that connects points $x_2$ and $x_5$ encode values of $x_6$ and $x_7$.

These visual representations are not pure ILC representations with a single baseline anymore, but rather ILC based on it, therefore we will call this type of coordinates **In-Line Based Coordinates** (ILBC).

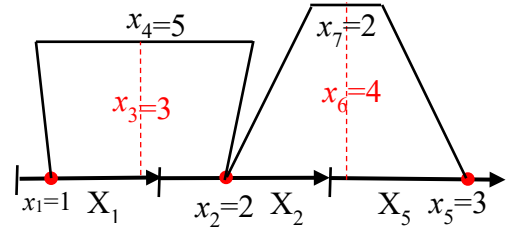

(a) $x_3$, $x_4$, $x_6$ and $x_7$ encoded by the height and width of the link lines that connect $(x_1,x_2)$ and $(x_2,x_5)$.

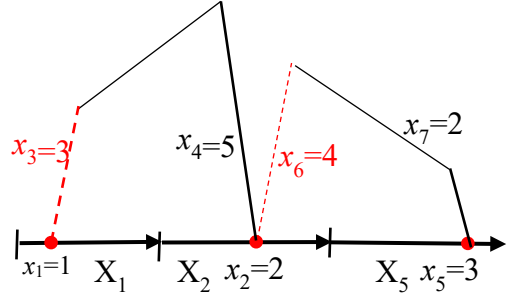

(b) $x_3,x_4,x_6$ and $x_7$ encoded by of length of sides of the link lines that connect $(x_1,x_2)$ and $(x_2,x_5)$.

Fig. 3. 7-D point **x** = $(x_1,x_2,x_3,x_4,x_5,x_6,x_7)$ = (1,2,3,5,3,4,2) in ILBC.

Fig. 4a simplifies Fig. 3b, by making sides **vertical**, and Figure 4b simplifies Fig. 4a further by removing vertical lines, which go down to the baseline and keeping only solid lines.

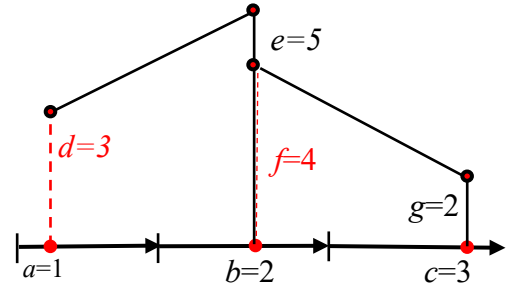

(a) Vertical simplification of Fig. 3b

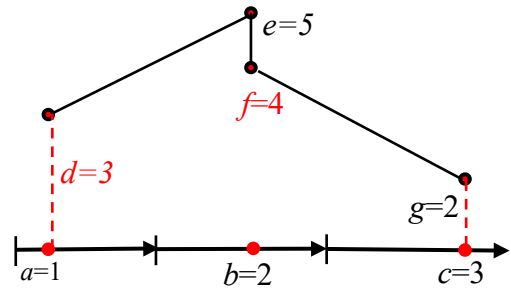

(b) Minimized representation

Fig. 4. 7-D point **x** = $(a,b,c,d,e,f,g)$ = (1,2,3,5,3,4,2) in ILC2 with vertical sides.

Also, a more generic simplified notation is used in Fig. 4 with attributes named from *a* to *g*, because any of the coordinates $\{x_i\}$ can be put to the baseline or link lines. Fig. 4 allows a full restoration of all 7 values and requires for them only 4 nodes and 3 edges, while parallel coordinates require 7 nodes and 6 edges.

The visual representation in Fig. 4 with vertical sides can be interpreted as follows. All vertical values *a-g* are located on respective coordinates A-G, which are vertically collocated on what is commonly known as Cartesian Y coordinate, while the ILC baseline occupies the Cartesian X coordinate. Thus, ILBC in Fig. 4 is a **combination of two ILCs – horizontal and vertical**. We will denote such coordinates **ILC2**.

Next, we compare ILC2 with Shifted Paired Coordinates (SPC) [11] on the same 7-D point shown in Fig. 5. In SPC in Fig. 5a, coordinates *d, e, f, g* are also vertical, but start at the origins of individual horizontal coordinates, which are paired. SPC also requires 4 nodes and 3 edges that are longer than in ILBC in Fig 4b. In ILC and ILBC above, the location of all coordinates on the horizontal baseline is fixed with their values located on this baseline.

Figs. 6 and 7 show Wisconsin Breast Cancer (WBC) data of two classes in partial ILC2 defined above. Drawing classes "mirrored" in Fig. 6 allows to compare and see the difference and similarities of patterns of two classes without their occlusion. Fig. 7 shows much better separation of WBC classes in fully dynamic ILC2.

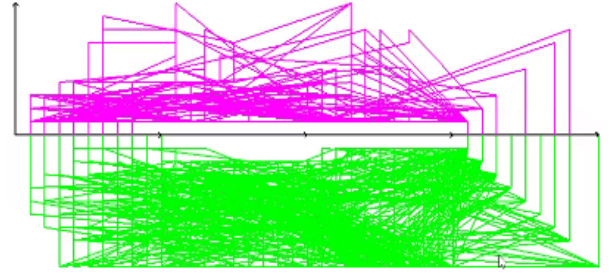

Fig.6. Wisconsin Breast Cancer data of two classes "mirrored" in partial ILBC.

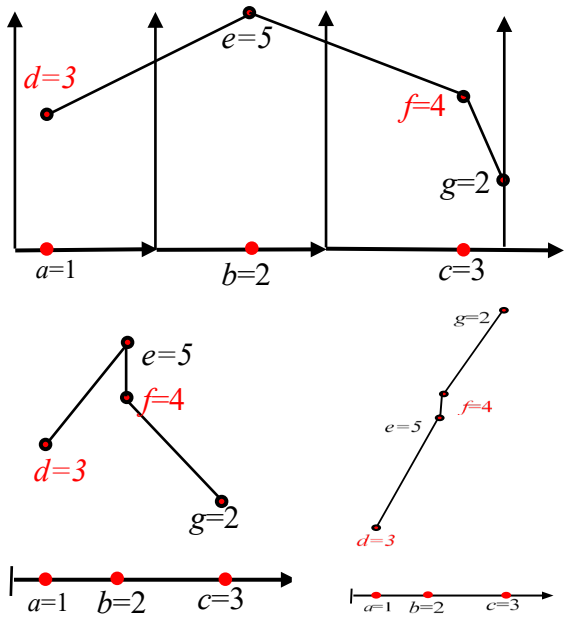

Fig. 5. 7-D point **x** = (*a,b,c,d*,e,*f,g*) = (1,2,3,5,3,4,2) in SPC (a) and ILBC partial dynamic (b) and full dynamic (c).

It is called a **static mapping** [11]. In the **dynamic mapping** of the given n-D point **x**, the location of the next value $x_{i+1}$ in its 2-D graph **x**\* depends on the location and prior values. It is a common concept for all General Line Coordinates [11], not only ILC and ILBC.

Fig. 5b shows the same 7-D point. Here the coordinate B starts at point *a*=2 of coordinate A with value *b*=2 located at distance 2 from point *a*=1. Respectively, coordinate C starts at point *b*=2 with point *c*=3 located at the distance 3 from *b*=2. The respective vertical coordinates D, E, F, and G start at the origin of the horizontal baseline. Thus, all of them are collocated and static. As we see here horizontal coordinates are dynamic, but vertical are static, therefore, we call this ILBC **partial dynamic ILC2**. Fig. 5c shows a **full dynamic ILC2** where vertical coordinates are dynamic in the same ways as horizontal coordinates, where the location of *e, f* and *g* points depends on the location of their prior points.

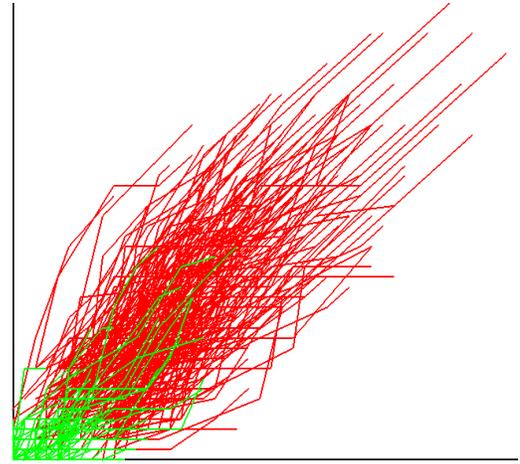

Fig. 7 WBC data in fully dynamic ILC2

The general formula to locate pairs ($x_i$, $x_j$) in **partial dynamic IL2** is given by mapping *L* as follows:

$$L(x_i, x_j) = (x_1+x_3+\ldots+ x_i, x_j).$$

Respectively the general formula to locate pairs ($x_i$, $x_j$) **fully dynamic ILBC** is given by mapping *L* as follows:

$$L(x_i, x_j) = (x_1+x_3+\ldots+ x_i, x_2+x_4+\ldots+ x_j).$$

Next, we introduce **weighted dynamic ILBC** with is given by mapping $L_w$ as follows:

$$L_w(x_i, x_j) = (w_1x_1+ w_3x_3+\ldots+ w_ix_i, w_2x_2+ w_4x_4+\ldots+ w_jx_j)$$

where W={$w_i$} is a set of weights assigned to coordinates.

III. CLASSIFICATION AND REGRESSION ALGORITHMS WITH IN-LINE COORDINATES

*A. Box classification algorithm*

The main idea of the **box classification algorithm**, which in short, we will call the **box algorithm** is finding a good box with high purity and a large number of cases. The algorithm records this box and removes all cases, which are in this box, and repeat the process of finding other good boxes in the remaining cases.

It continues until it cannot find more good boxes. This process is interactive and partially automated. Automation includes computing parameters of the candidates for the good boxes.

The box algorithm operates on n-D data visualized in ILBC in the following major steps.

Step 1: Search/discover "good" boxes $B_i$ in these visualizations that cover cases of a single class $C_i$ as pure as possible (max of purity and large number of cases covered). The search of boxes and rules is a sequential **hierarchical process** for each class. Here it is assumed that box B *covers* the n-D point **x** if *one or more nodes* of graph **x*** of **x** is in box B. This definition leads to simpler interpretation of rules based on such boxes, than an alternative definition, which only requires that an edge to graph x* crosses the box B.

Step 2: Form basic box classification rules with the discovered boxes:

$$R_i: \text{if } \mathbf{x} \in B_i \Rightarrow \mathbf{x} \in \text{Class } C_i \quad (1)$$

The general format of the rules is

$$R_i: \text{if } \mathbf{x} \in B_i \ \& \ \mathbf{x} \notin (B_m \cup B_p \cup \ldots \cup B_t) \Rightarrow \mathbf{x} \in \text{Class } C_i \quad (2)$$

Here $B_i$ is a current "good" box, and other boxes are prior "good" boxes with cases from these boxes removed before $B_i$ is searched.

Step 3: Test box classification rules on independent data.

Step 4: Pruning a set of discovered rules to decrease overfitting.

Fig. 8 illustrate basic boxes and box rules discovered for WBC data, where

$R_1$: if $\mathbf{x} \in B_1 \Rightarrow \mathbf{x} \in$ Class Green, $R_2$: if $\mathbf{x} \in B_2 \Rightarrow \mathbf{x} \in$ Class Red,

Box $B_1$ contains 292 green cases and box $B_2$ contains 12 red cases.

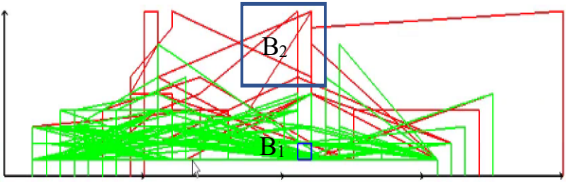

Fig. 8. Examples of boxes discovered in WBC data.

The main substeps of Step 1 (discovering boxes) are:
1.1. Divide the ILC area where graphs of all n-D cases are located to boxes by using a grid. The boxes are typically quite small and of a fixed size. The details of this process will be elaborated later.
1.2. If the number of boxes and the number of cases are relatively small, compute purity of each box (number of cases of each class in the box, and ratio of the cases of the dominant class to the number of cases of all other classes), else use optimization and heuristic algorithms such as Genetic Algorithms (GAs).
1.3. For each class create a list of boxes where this class dominates.
1.4. For a given class pick up all pure boxes, if a set of pure boxes is not empty.
1.5. Create a classification rule with these boxes.
1.6. Exclude all cases that are in these boxes.
1.7. Conduct 1.4-1.6 for all classes.
1.8. Create boxes from remaining cases.
1.9. For each class order boxes from 1.8 in the purity descending order.
1.10. For each class order boxes from 1.8 in the number of cases descending order.
1.11. Create a joint descending order of boxes from 1.9 and 1.10 (currently an interactive process).
1.12. For a given class pick up the top box from 1.11 and create a rule with it.
1.13. Exclude cases from 1.11 and repeat steps 1.8-1.12.

The pruning step is to deal with many "mini" boxes that contain too few cases with a low level of generalization, and likely data overfitting and memorization. This problem is also known for decision trees. Without control the number of layers and the number of cases in each terminal node, decision trees produce many terminal nodes with only few cases in each of them. Such pruning removes overfitting but leads to decreasing the accuracy of classification by decision trees.

The step 4 of the box algorithm employs a version of this pruning approach: (a) associate "mini" boxes with the larger boxes or (b) refuse to predict cases that belong to "mini" boxes. The interactivity of the Box algorithm has an advantage of allowing the end-users to observe "mini" boxes and decide to follow (a) or (b) to deal with them.

B. *3.2. Linear classification and regression algorithm*

Design of rules based on boxes have limitations. One of them is *locality* of each box, i.e., the box covers only cases that are in this box. Typically, we need several boxes to cover all data, while a single linear classifier can cover all data if data are linearly separable. The goal of this section is proposing an analog of linear classifier in ILC. Fig. 9 illustrates the proposed approach. First, we built a black line which is used to project all cases of both classes to this line. If the projected endpoints of cases of one class *C* mostly concentrate on the one side of the discrimination blue line, then we say that a linear discrimination model *M* is discovered:

$$M(\mathbf{x}) > T \Rightarrow \mathbf{x} \in C \quad (3)$$

Another, more common, linear discrimination model for two classes C and *Q* is:

$$M(\mathbf{x}) > T \Rightarrow \mathbf{x} \in C \text{ else } \mathbf{x} \in Q \quad (4)$$

The model (3) only tells about points, where $M(\mathbf{x}) > T$, but nothing about points, where $M(\mathbf{x}) \leq T$. This is the situation in Fig. 9 for red class above the blue line. This means that if we cannot build a single model (4) we need to build several models like (3) that may require different black lines, where endpoints are projected as shown in Fig 9b. Moreover, we can relax a requirement that only endpoints are projected. We can project some intermediate nodes $\mathbf{x}_k$ of graph of **x*** getting models like

$$M(\mathbf{x}_k) > T \Rightarrow \mathbf{x} \in C \quad (5)$$

$$M_1(\mathbf{x}_k) > T_k \ \& \ M_2(\mathbf{x}_u) > T_u \Rightarrow \mathbf{x} \in C \quad (6)$$

Those intermediate points can be found by the Box algorithm presented above. Below an ILC classification and regression algorithms are presented.

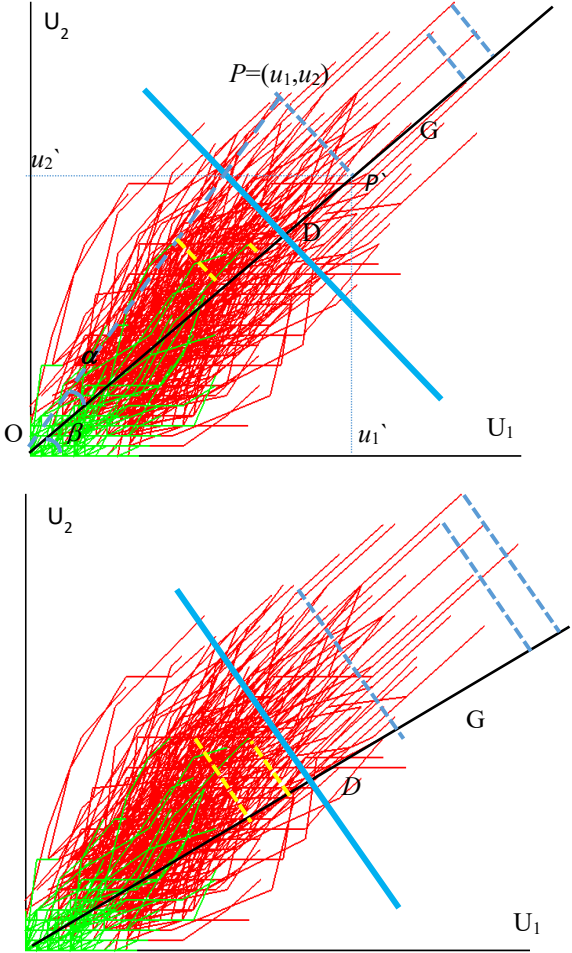

Fig. 9. Classification and regression with projection line at different angles in full dynamic ILP2.

IV. CASE STUDY FOR BOX ALGORITHM IN PARTIAL DYNAMIC ILC2

This section presents the results of the computational experiment for discovering classification rules for WBC data encoded in partial dynamic ILC2 using the Box algorithm. We analyze rules, and the ways how these rules can be simplified and pruned. The discovered 13 pure boxes are presented in Table 1. These boxes cover all WBC cases. Values $x_1$, $x_2$, $y_1$, $y_2$ identify left, right, bottom and top corners of the box within ILC2.

Table 1. Discovered boxes in partial dynamic ILC2.

| Box | $x_1, x_2, y_1, y_2$ | cases | Box | $x_1, x_2, y_1, y_2$ | cases |
|---|---|---|---|---|---|
| $B_1$ | 15,20.5,1,1.5 | **382** | $B_2$ | 23.5,39.5,8.5,10 | **166** |
| $B_3$ | 1,3.5,0.5,2 | 28 | $B_4$ | 20,22.5,6,6.5 | 26 |
| $B_5$ | 9.5,10.5,6.5 | 14 | $B_6$ | 16,21,0.5,2 | 18 |
| $B_7$ | 17.5,18.5,3,3.5 | 23 | $B_8$ | 14.5,17,2.5,3 | 7 |
| $B_9$ | 28.5,29,2.5,3.5 | 4 | $B_{10}$ | 17.5,18.5,3,3.5 | 10 |
| $B_{11}$ | 14.5,15,5.5,6 | 4 | $B_{12}$ | 26.5,27,7,7.5 | 1 |
| $B_{13}$ | 28,28.5,0.5,9.5 | 10 | | | |

Table 2 presents the rules constructed from these boxes, in the hierarchical process of the Box algorithm, which was described above. The benign class is denoted as B and is drawn as Green with letter G used to identify this class in Table 2. Respectively, the class malignant is denoted by M and Red (R) for short.

Table 2. Rules $R_1$-$R_{13}$ with precision P=100%.

| *Benign, B (green, G) class rules* |
|---|
| $R_1$: $\mathbf{x} \in B_1 \Rightarrow \mathbf{x} \in G$ (382 cases) |
| $R_3$: $\mathbf{x} \in B_3 \Rightarrow \mathbf{x} \in G$ (28 cases) |
| $R_6$: $\mathbf{x} \in B_6$ & $\mathbf{x} \notin B_2 \cup B_4 \cup B_5 \Rightarrow \mathbf{x} \in G$ (18 cases) |
| $R_8$: $\mathbf{x} \in B_8$ & $\mathbf{x} \notin B_2 \cup B_4 \cup B_5 \cup B_7 \cup B_{10} \cup B_{13} \Rightarrow \mathbf{x} \in G$ (7 cases) |
| $R_{8s}$: $\mathbf{x} \in B_8$ & $\mathbf{x} \notin B_2 \cup B_4 \cup B_5 \cup B_7 \cup B_{10} \Rightarrow \mathbf{x} \in G$ (7 cases) |
| (no $B_{13}$ specific cases, the rule is simplified) |
| $R_9$: $\mathbf{x} \in B_9$ & $\mathbf{x} \notin B_2 \cup B_4 \cup B_5 \cup B_7 \cup B_{10} \cup B_{13} \Rightarrow \mathbf{x} \in G$ (4 cases) |
| $R_{9s}$: $\mathbf{x} \in B_9$ & $\mathbf{x} \notin B_2 \cup B_4 \cup B_5 \cup B_7 \Rightarrow \mathbf{x} \in G$ (4 cases) |
| (no $B_{10}$, $B_{13}$ specific cases, the rule is simplified) |
| $R_{11}$: $\mathbf{x} \in B_{11}$ & $\mathbf{x} \notin B_2 \cup B_4 \cup B_5 \Rightarrow \mathbf{x} \in G$ (4 cases) |
| $R_{12}$: $\mathbf{x} \in B_{12}$ & $\mathbf{x} \notin B_2 \cup B_4 \cup B_5 \cup B_7 \cup B_{10} \Rightarrow \mathbf{x} \in G$ (1 case) |
| *Malignant, M (red, R) class rules* |
| $R_2$: $\mathbf{x} \in B_2 \Rightarrow \mathbf{x} \in R$ (166 cases) |
| $R_4$: $\mathbf{x} \in B_4 \Rightarrow \mathbf{x} \in R$ (26 cases) |
| $R_5$: $\mathbf{x} \in B_5$ & $\mathbf{x} \notin B_1 \cup B_3 \Rightarrow \mathbf{x} \in R$ (14 cases) |
| $R_{5s}$: if $x \in B_5$ & $x \notin B_1$ ) $\Rightarrow \mathbf{x} \in R$ (14 cases) |
| $R_7$: $\mathbf{x} \in B_7$ & $\mathbf{x} \notin B_1 \cup B_3 \cup B_6 \Rightarrow \mathbf{x} \in R$ (13 cases) |
| ($B_3$ cases belong to box $B_1$ thus the rules are simplified) |
| $R_{7s}$: $\mathbf{x} \in B_7$ & $\mathbf{x} \notin B_3 \cup B_6 \Rightarrow \mathbf{x} \in R$ (13 cases) |
| $R_{10}$: $\mathbf{x} \in B_{10}$ & $\mathbf{x} \notin B_3 \cup B_6 \cup B_8 \cup B_9 \Rightarrow \mathbf{x} \in R$ (10 cases) |
| $R_{13}$: $\mathbf{x} \in B_{11}$ & $\mathbf{x} \notin B_1 \cup B_3 \cup B_6 \cup B_8 \cup B_9 \cup B_{11} \cup B_{12} \Rightarrow \mathbf{x} \in R$ (10 cases) |

The boxes and rules in Tables 1 and 2 cover all 444 B cases and 239 M. Boxes $B_1$-$B_4$ and respective rules $R_1$-$R_4$ cover most of the cases (602 cases) with 100% precision without any misclassified cases. This means that 88.14% of all cases are classified by simplest rules with a single box without any other boxes involved. The other rules involve "*negated*" boxes requiring that the case does not belong to these boxes to satisfy the rule. Some rules in Table 2 such as rule $R_8$ have simplified forms too with reduced "negated" boxes, because all cases of some boxes are covered by other boxes in these rules.

Table 2 shows that class G has more rules/boxes with smaller coverage (4 rules that cover from 1 to 7 cases with total 16 cases covered by these rules). We will call these boxes "*mini*" *boxes*. In contrast, class R has no rules and boxes with such small coverage. Its four rules with smaller coverage include from 10 to 14 cases with 47 total cases. It means a better generalization for R class, than for G class, on these rules.

When we analyze the rules with remaining larger coverage, the situation is the opposite. The first three G rules cover 428 cases (96.4% of G cases), while the first two R rules (rules $R_2$ and $R_4$) cover 192 cases, 80.3% of R cases). The rules that cover a small number of cases are more complex. They include 2 to 7 "negated" boxes. This is rather a memorization of cases, than their generalization. Such complex rules are needed for a small number of cases.

We are not the first ones, who observed that often pairs of attributes are sufficient for classifying *most of the cases*, and in a medical domain specifically [2]. It means that domain experts correctly captured/engineered a few critical pairs of mutually dependent attributes. This can indicate superior human abilities to generate informative features manually. While, deep learning algorithms can automatically discover informative features, often they are complex, and it is challenging to interpret and explain them.

A comparable decision tree built on the 90% of the same data is 100% accurate too, but has multiple terminal nodes with few cases in each of them. It rather overfits and memorizes data in such terminal nodes. This tree has a total of 57 nodes, 29 terminal nodes with 20 of these terminal nodes contain 7 or less cases [8]. In contrast the box algorithm produced 13 boxes/rules, and only 4 of them contain 7 or less cases on all WBC data.

Figs. 10-20 illustrate all boxes showing the cases, which are in these boxes. In addition, Figs. 11-20 show also the cases of other colors, which are removed before discovering a given box by requiring not to belong to a set of prior boxes, listed in the rules in Table 2.

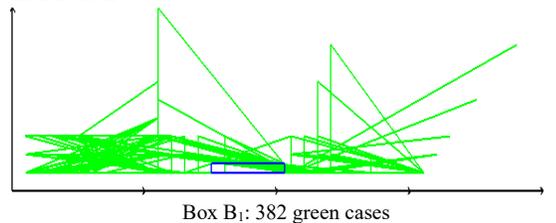

Box $B_1$: 382 green cases

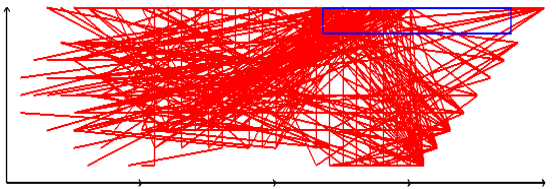

Box $B_2$: 166 red cases
Fig. 10. Boxes $B_1$, $B_2$.

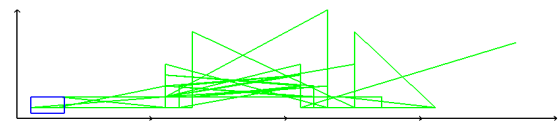

Box $B_3$: covers 28 green cases.

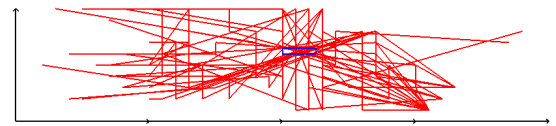

$B_4$: covers 26 red cases.
Fig. 11. Boxes $B_3$, $B_4$.

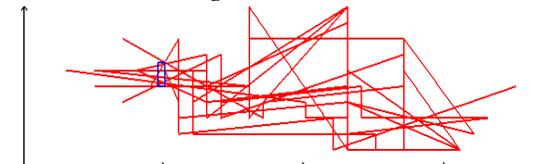

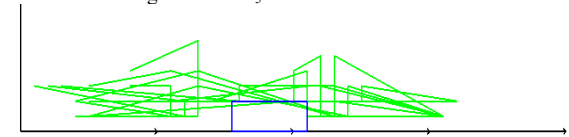

Fig. 12. Box $B_5$: covers 14 red cases.

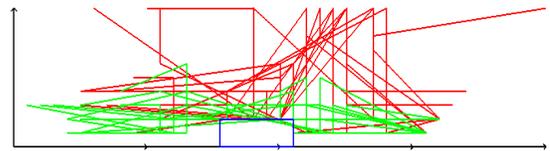

Fig. 13. Box $B_6$: 18 green cases.

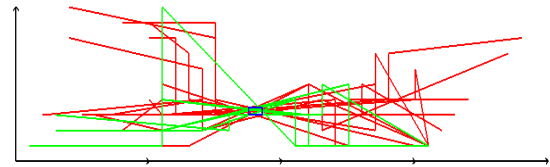

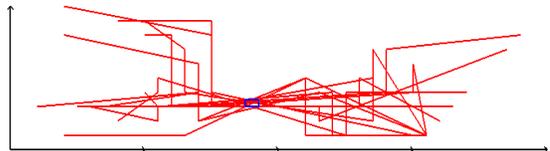

Fig. 14. Box $B_7$: 23 red cases.

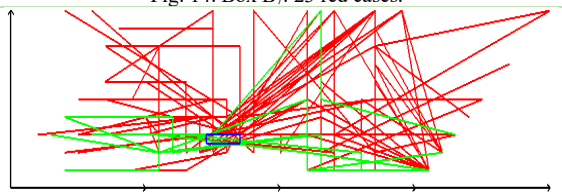

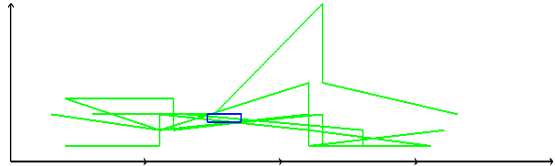

Fig. 15. Box $B_8$: 7 green cases.

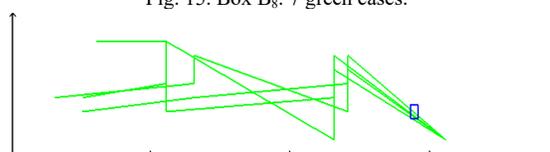

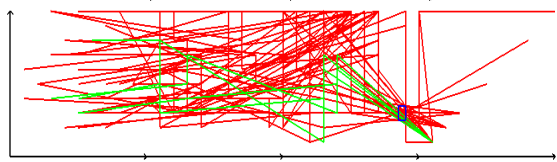

Fig. 16. Box $B_9$: 4 green cases.

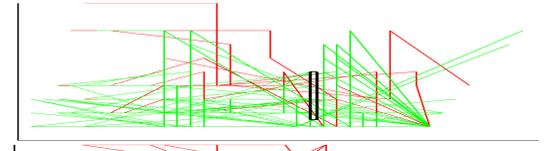

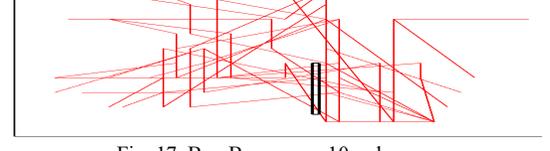

Fig. 17. Box $B_{10}$: covers 10 red cases.

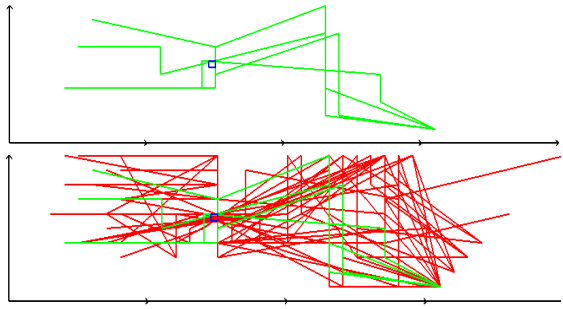
Fig 18. Box $B_{11}$: 4 green cases.

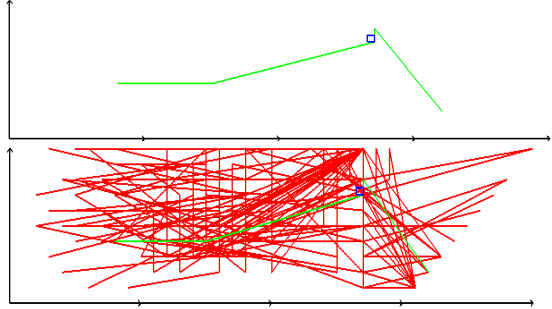
Fig. 19. Box $B_{12}$: 1 green case

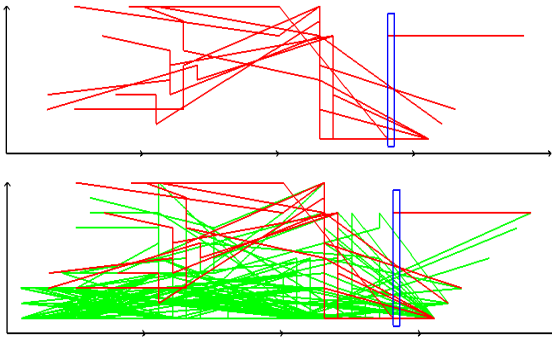
Fig. 20. Box B13: 10 red cases.

**Pruning**. Boxes $B_9$, $B_{11}$ with 4 green cases each and box $B_{12}$ with 1 green case can be used to prune the set of rules by creating modified rules:

$R_{9M}$: $\mathbf{x} \in B_9 \Rightarrow \mathbf{x} \in R$ (47 red /4 green)
$R_{11M}$: $\mathbf{x} \in B_{11} \Rightarrow \mathbf{x} \in R$ (28 red /4 green)
$R_{12M}$: $\mathbf{x} \in B_{12} \Rightarrow \mathbf{x} \in R$ (52 red /1 green)

These rules have low error rate. The end user can explore them and accept if this error rate is tolerable.

**Joining rules**. The next task is decreasing the number of rules, which is demonstrated on 13 rules shown above. The proposed approach joins rules by combining them, including the use of the else condition. In contrast with the pruning, this process *does not introduce any error*. The result is shown in Table 3.

The steps of the **Rule Joining (RJ) algorithm** are:
Step 1: Combine rules with a single rectangle of a given class.
Example: $R_{1,3}$: $x \in B_1 \cup B_3 \Rightarrow x \in G$ (410 cases)
Step 2: Find rules in the opposite class that are conditioned by rectangles used in Step 1. Example: Rule $R_5$ is conditioned by
$\mathbf{x} \notin B_1 \cup B_3$.

Step 3: Combine rules from Steps 1 and 2. Example: $R_{1,3,5}$: $x \in B_1 \cup B_3 \Rightarrow x \in G$ (else $x \in B_5 \Rightarrow x \in R$) (428 cases)

Here Rule $R_5$ covers only 14 cases, which can be viewed as a potential overfitting, while rule $R_{1,3,5}$ covers 428 cases. The else condition makes $R_5$ a part of the larger rule.

Table 3. Rules after joining.

| |
|---|
| *Expanded benign (green, G) class rules* |
| $R_{1,3}$: $x \in B_1 \cup B_3 \Rightarrow x \in G$ (410 cases) |
| $R_{1,3,5}$: $x \in B_1 \cup B_3 \Rightarrow x \in G$ (else $x \in B_5 \Rightarrow x \in R$) (424 cases) |
| $R_{8s,9s}$: $x \in B_8 \cup B_9$ & $x \notin B_2 \cup B_4 \cup B_5 \cup B_7 \Rightarrow x \in G$ (11 cases) |
| $R_{11,12}$: $x \in B_{11} \cup B_{12}$ & $x \notin B_2 \cup B_4 \cup B_5 \Rightarrow x \in G$ (5 cases) |
| *Expanded malignant (red, R) class rules* |
| $R_{2,4}$: $x \in B_2 \cup B_4 \Rightarrow x \in R$ (192 cases) |
| $R_{2,4,6}$: $x \in B_2 \cup B_4 \Rightarrow x \in R$ (else $x \in B_6 \Rightarrow x \in G$) (210 cases) |
| $R_{7s}$: $x \in B_7$ & $x \notin B_3 \cup B_6 \Rightarrow x \in R$ (13 cases) |
| $R_{2,4,8}$: $x \in B_2 \cup B_4 \Rightarrow x \in R$ (else $x \in B_8$ & $x \notin B_5 \cup B_7 \Rightarrow x \in G$) (199 cases) |
| $R_{2,4,6,8}$: if $x \in B_2 \cup B_4 \Rightarrow x \in R$ (else $x \in B_6 \cup B_8$ & $x \notin B_5 \cup B_7 \Rightarrow x \in G$) (217 cases) |
| $R_{10}$: $x \in B_{10}$ & $x \notin B_3 \cup B_6 \cup B_8 \cup B_9 \Rightarrow x \in R$ (10 cases) |
| $R_{13}$: $x \in B_{13}$ & $x \notin B_1 \cup B_3 \cup B_6 \cup B_8 \cup B_9 \cup B_{11} \cup B_{12} \Rightarrow x \in R$ (10 cases) |

The analysis of rules in Table 3 shows that 7 rules $R_{1,3,5}$, $R_{8s,9s}$, $R_{11,12}$, $R_{2,4,6,8}$, $R_{7s}$, $R_{10}$, and $R_{13}$ are equivalent to 13 original rules. Here rules $R_{8s,9s}$, $R_{11,12}$, $R_{10}$, and $R_{13}$ cover 10 or less cases with total 36 cases (16 green, and 20 red). Excluding these rules and, respectively refusing to classify cases, which satisfy them, will eliminate potential overfitting.

*A. K-fold cross validation*

So far, the case study was conducted on the whole WBC dataset. What will be the accuracy of the box algorithm in k-fold cross validation (k-fold CV) on these data? We answer this question in a *non-traditional way*. It is an attempt to find the *worst and best-case estimates* for 10-fold cross validation as follows. The formal concept of worst-case cross validation estimate, based on the Shannon function, was introduced in [9]. The motivation of getting the worst-case estimates is coming from the fact that *k*-fold CV only tests a small fraction of splits of data into training and validation sets, giving potentially an *inflated average estimate*, which can be misleading in life-critical applications such as cancer diagnostics.

First, consider 10-fold CV with a validation fold, which includes all 16 cases that are in "mini" boxes $R_8$-$R_9$ and $R_{11}$-$R_{12}$. The training data in the remaining 9 folds do not contain these cases. Thus, these "mini" boxes will not be discovered by the box algorithm. Assume that this algorithm discovered all other boxes on training data that contain 90% of all WBC.

How the Box algorithm will classify cases from these "mini" boxes? There are two options: (1) refusal and (2) make an error by using modified rules like $R_{12M}$ because they misclassify these cases. In 10-fold CV, we have a training – validation split of cases: 615 – 68. In the worst validation fold 16 cases are misclassified with 76.47% accuracy. All other folds do not have any misclassified cases (100% accuracy), because we assumed that all other boxes are discovered. The average accuracy in all 10-folds will be average of 76.47% and 100% taken nine times – 97.65%. If we refuse to classify these

16 cases, then the precision will be 100%. Both situations (1) and (2) can be considered as the *best-case* estimates.

Next, we relax the assumption that all non-mini boxes $B_i$ are discovered exactly. Assume that smaller boxes $B_i`$ are discovered, which do not include all 52 remaining cases in the worst fold. Then the accuracy in the worst fold will be 0% with 100% accuracy in all 9 other folds with the average accuracy of 90%. This is a *worst-case* estimate. The average estimate will be between found best and worst estimates.

A more detailed analysis is also possible, which involves visual analysis. Let's consider the largest box $B_1$ with 382 green cases. In stratified 10-fold CV, only 38 cases from this box are left for the validation fold. Fig. 20 shows the location of this box. The visual analysis allows us to identify and count lines which are at the edge of the box. If that number is less than 38 cases, then we know that misclassifying all 38 cases is impossible, when we learn $B_1$ with a subset of data.

## V. CONCLUSION

This paper proposed a new approach for machine learning based on 2-dimensional space (2-D ML) in the Inline Coordinates. It is shown that this is a full machine learning approach, which does not require to deal with n-dimensional data in n-dimensional space. It works with n-dimensional data in 2-dimensional space without any loss of information. It is done with inline based coordinates in different modifications that we defined including both static and dynamic ones. Then classification and regression algorithms based on these inline coordinates were introduced. A successful case study based on WBC data demonstrated the feasibility of the approach.

This work expands a *whole new area of full 2-D machine learning with a new methodology*. It has advantages of abilities to actively include the end-users into the machine learning discovering process and justification of models. It also allows to provide interpretable ML models. For the static coordinates, this follows directly from their definition, because all boxes are discovered in original interpretable attributes. For the dynamic coordinates the situation is a more complex, because the relation of the coordinates is expressed by location of the next point depending on the location of previous one. It is given by two linear functions $F_1$ and $F_2$ of values of all previous coordinates. Respectively, each box is defined by upper and lower limits for these functions.

Often it is stated [15] that linear functions are directly interpretable because the contribution of each attribute is easily expressed by their coefficients. A more elaborated approach is presented in [7] where a step function is used to interpret linear dependence. Thus, in both static and dynamic situations we have an interpretable model for the Inline-based Coordinates presented here. The future work is strengthening algorithms for getting higher accuracy, and widening the set of possible patterns, which can be discovered in Inline Coordinates. Here we only experimented with boxes. An alternative approach with a linear function for classification and regression, which we also described in this paper, needs experimental studies and deeper analysis. Other future work is dealing with the large number of cases using multithreading and the GPU.